\documentclass[conference]{IEEEtran}
\IEEEoverridecommandlockouts

\usepackage{cite}
\usepackage{amsmath,amssymb,amsfonts}
\usepackage{graphicx}
\usepackage{textcomp}
\usepackage{xcolor}
\usepackage{hyperref}
\usepackage{booktabs}
\usepackage{array}
\usepackage{multirow}
\usepackage{url}

\begin{document}

\title{A Deterministic Agentic Workflow for HS Tariff Classification: Multi-Dimensional Rule Reasoning with Interpretable Decisions}

\author{
\IEEEauthorblockN{Yu Zhang\IEEEauthorrefmark{1},
Dongjiang Zhuang\IEEEauthorrefmark{5},
Qu Zhou\IEEEauthorrefmark{3},
Zheng Huang\IEEEauthorrefmark{4},
Junhe Wu\IEEEauthorrefmark{2},
Jing Cao\IEEEauthorrefmark{2},
Kai Chen\IEEEauthorrefmark{1}\thanks{Corresponding author: Kai Chen (kchen@sjtu.edu.cn).}}
\IEEEauthorblockA{\IEEEauthorrefmark{1}School of Information and Electronic Engineering,
Shanghai Jiao Tong University, Shanghai, China}
\IEEEauthorblockA{\IEEEauthorrefmark{2}Customs National Supervision Bureau for Duty Collection (Shanghai), \\
General Administration of Customs of the P.R.C. (GACC), Shanghai, China}
\IEEEauthorblockA{\IEEEauthorrefmark{3}Nanjing Jiyun Information Technology Co., Ltd., Nanjing, China}
\IEEEauthorblockA{\IEEEauthorrefmark{4}School of Computer Science,
Shanghai Jiao Tong University, Shanghai, China}
\IEEEauthorblockA{\IEEEauthorrefmark{5}Department of Science and Technology (Shanghai), \\
General Administration of Customs of the P.R.C. (GACC), Shanghai, China}
}

\maketitle

\begin{abstract}
Harmonized System (HS) tariff classification is a high-stakes, expert-level
task in which a free-form product description must be mapped to a specific
six- or eight-digit code under the General Interpretive Rules (GIR), section
notes, chapter notes, and Explanatory Notes. The difficulty lies not in
knowledge volume but in \emph{multi-dimensional rule reasoning}: a correct
classification must satisfy competing priority rules along several axes
simultaneously, including material, form, function, essential character,
the part--versus--whole boundary, and specific listing versus residual
headings. End-to-end prompting of large language models fails
characteristically by resolving one axis while ignoring the priority
constraints on the others. We present a \emph{deterministic agentic
workflow} in contrast to self-planning agents: the control flow is fixed,
language model calls are confined to narrow stages, and reflection and
verification are retained as local mechanisms. This design yields
interpretability by construction---each decision is decomposed into
stage-wise structured outputs with verbatim citation of the chapter or
section notes that bear on it. The architecture combines offline
knowledge-engineering of the Chinese HS tariff with an online six-stage
pipeline. Evaluated on HSCodeComp at the six-digit level, the workflow
reaches 75.0\% top-1 and 91.5\% top-3 at four digits, and 64.2\% top-1
and 78.3\% top-3 at six digits with Qwen3.6-plus; an open-weight
Qwen3.6-27B-FP8 backbone in non-thinking mode achieves 84.2\% four-digit
and 77.4\% six-digit top-1 agreement with the frontier model. A two-stage
manual audit of 226 six-digit disagreements suggests that a non-trivial
fraction of HSCodeComp ground-truth labels may deviate from HS general
rules; full adjudication records are released in the appendix as
preliminary findings for community review.
\end{abstract}


\section{Introduction}

The Harmonized Commodity Description and Coding System, commonly known as the
Harmonized System (HS), is the global nomenclature for traded goods,
established and maintained by the World Customs Organization (WCO) and
implemented in more than 200 customs administrations. The system organizes
roughly 5{,}000 commodity groups under a hierarchical structure of 21
sections, 99 chapters, approximately 1{,}200 four-digit headings, and several
thousand six-digit subheadings; many national administrations, including
China's General Administration of Customs (GACC) and the United States
International Trade Commission, further extend the six-digit international
codes to eight or ten national digits for duty and statistical purposes.
Classification of a given product to a specific code is the legal basis on
which import duties, value-added tax, anti-dumping levies, dual-use export
controls, and a wide range of regulatory permissions are assigned. The
economic stakes are correspondingly high: misclassification can trigger
post-entry duty adjustments, shipment holds, civil penalties, and---in cross
border e-commerce---temporary suspensions of postal corridors. Classification
is performed every day at industrial scale by customs officers, brokers, and
trade-compliance teams; the WCO reports that more than one billion customs
declarations are processed worldwide each year.

The challenge of HS classification is, surprisingly, not one of knowledge
volume. The tariff itself is a finite, public document; the section notes,
chapter notes, and Explanatory Notes maintained by the WCO together with
national customs administrations occupy at most a few thousand pages. The
difficulty lies in what we call \emph{multi-dimensional rule reasoning}: a
correct classification must satisfy competing priority rules along several
axes simultaneously. At least six such axes recur across the tariff:
(i) \emph{material}, distinguishing for instance plastic from rubber, textile
from leather, or steel from aluminum; (ii) \emph{form and processing state},
distinguishing primary forms (granules, powders, liquids) from intermediate
forms (rods, profiles, monofilaments), from finished articles, from parts;
(iii) \emph{function and end-use}, distinguishing general-purpose articles
from those dedicated to a specific machine, vehicle, or medical procedure;
(iv) \emph{essential character} under GIR~3(b), which arbitrates the
classification of composite articles by identifying the component that gives
the article its essential character; (v) the \emph{part-versus-whole
boundary} under Section~XVI Note~2 and Section~XVII Note~2, which decides
whether a component is classified with the parent machine or as a separate
article in its own right; and (vi) \emph{specific listing versus residual
headings} under GIR~1 and GIR~3(a), which mandates that named goods take
priority over generic basket headings.

These axes are not independent. The chapter notes themselves embed
cross-references and exclusions that re-route a candidate classification
from one chapter to another; for example, the chapter notes of Chapter~39
specify a strict form-priority architecture in which a finished article in
plastic (e.g., a 1.75\,mm-diameter monofilament made of PLA) must be
classified by its form (heading 39.16, plastic monofilaments) rather than
by its underlying polymer (subheading 3907.70, polylactic acid). A
practitioner classifying the same item without this rule will land on the
polymer subheading---a defensible choice on the material axis but an
incorrect one once form priority is taken into account.

End-to-end prompting of large language models (LLMs) on HS classification
exhibits exactly this failure pattern. The model resolves one axis
correctly---most often material, occasionally function---and produces a
fluent, rule-citing rationale that nonetheless ignores the priority
constraints on the other axes. The problem is compounded by the fact that
six- and eight-digit codes are valid-looking integer strings: a model that
has not been grounded in the actual tariff index will readily fabricate
codes that do not exist in the schedule at all. Chain-of-thought prompting
alone does not fix this. The underlying problem is not insufficient
reasoning but the absence of the right slice of the corpus from the model's
context, together with the absence of any structural commitment about
\emph{where} in the corpus the next constraint should be loaded.

We report on a production-grade system for Chinese HS classification
designed around this observation. Rather than treat the task as a single
generation problem, we decompose it into a \emph{deterministic agentic
workflow}: a directed sequence of narrow, retrieval-grounded LLM stages
organized along the hierarchical structure of the tariff itself (chapter,
heading, subheading), with each stage emitting a structured, individually
testable artifact. Reflection, verification, and recovery are retained as
local mechanisms within stages, but the control flow itself is fixed by
engineers rather than discovered by the model. The design has two practical
consequences. First, interpretability arises by construction: every
classification decision is decomposed into stage-wise structured outputs
with verbatim citation of the chapter or section notes that bear on the
decision, allowing customs officers and declarants to audit each step
independently. Second, the workflow is moderate in its compute footprint: an
open-weight 27B-class backbone in non-thinking mode produces predictions
closely aligned with a frontier model on the same workflow, indicating that
the architecture does not rest on the reasoning capacity of any single
state-of-the-art LLM.

The contributions of this paper are as follows. We articulate
multi-dimensional rule reasoning as the characteristic difficulty of HS
classification and explain why end-to-end LLM prompting fails on it. We
present a deterministic agentic workflow---combining offline
knowledge-engineering of the Chinese HS tariff with an online six-stage
pipeline---and discuss its position relative to self-planning agents. We
evaluate on HSCodeComp~\cite{yang2025hscodecomp}, a recent expert-level
benchmark of 632 product entries, reporting six-digit accuracy as the
primary metric for reasons discussed in Section~V. We further conduct a
two-stage manual audit of disagreements between our predictions and
HSCodeComp ground-truth labels, the results of which suggest that a
non-trivial fraction of the labels may deviate from HS general rules; full
adjudication records are released as preliminary findings.

\section{Related Work}

\subsection{Datasets and benchmarks for HS classification}

Public datasets and benchmarks for HS or HTS classification have remained
scarce relative to the practical importance of the task. The most directly
relevant recent contribution is HSCodeComp~\cite{yang2025hscodecomp}, an
expert-level benchmark of 632 product entries drawn from a large-scale
e-commerce platform, annotated by multiple e-commerce domain experts at the
ten-digit level and spanning 27 chapters and 32 first-level categories.
HSCodeComp is the benchmark on which we evaluate in Section~V. ATLAS~%
\cite{yuvraj2025atlas}, released roughly contemporaneously, contributes a
larger-scale dataset of 18{,}731 rulings processed from the U.S.\ Customs
Rulings Online Search System (CROSS), with a held-out test split intended
for benchmarking large-model classification on U.S.\ HTS codes. An earlier
benchmark study by Judy~\cite{judy2024benchmark} evaluates four commercial
HTS classification tools---Zonos, Tarifflo, Avalara, and the WCO BACUDA
prototype---on 100 CROSS rulings, with metrics covering accuracy, speed,
rationality, and HTS-code alignment; this work is, to our knowledge, the
only published characterization of the commercial landscape on U.S.\ data.
Beyond these, the public dataset situation for non-U.S.\ tariff regimes
remains thin: we are not aware of an analogously sized, expert-annotated,
publicly available benchmark for the Chinese HS regime in the
English-language literature.

\subsection{Methods for HS classification}

Methods reported in the English-language literature fall into three broad
families. The first family treats the task as supervised text classification
or retrieval over heading text. The most developed entry in this family is
Lee et al.~\cite{lee2024explainable}, who, in collaboration with the Korea
Customs Service, propose an explainable decision-supporting model based on
Sentence-BERT with supervised contrastive learning and a multiple-negative-
ranking loss; the model retrieves precedent rulings as explanations and
reports 93.9\% top-3 accuracy on 925 challenging subheadings, validated by a
32-expert user study. Similar contrastive-learning approaches over trade
transactions have been explored for other national tariff regimes. These
systems are highly competitive within their training distribution but, by
construction, do not perform multi-step rule reasoning over GIR clauses or
chapter notes.

The second family fine-tunes large language models on tariff data.
ATLAS~\cite{yuvraj2025atlas} is the most recent representative work: a
LLaMA-3.3-70B model supervised-fine-tuned on the 18{,}731-ruling CROSS
training split, reporting 40\% ten-digit and 57.5\% six-digit accuracy on
held-out CROSS test data and outperforming several leading proprietary
reasoning models on the same task. Gholamian
et al.~\cite{gholamian2024llmrobust} study LLM in-context learning for
product classification under realistic data perturbations (abbreviations,
deletions), comparing flat, hierarchical-prompting, and few-shot LLM
configurations against fine-tuned Sentence-Transformer classifiers on
e-commerce taxonomies; they find LLMs to be substantially more robust to
noisy and abbreviated descriptions than supervised baselines, motivating
the use of LLMs over noisy small-importer descriptions.

The third family targets the \emph{agentic} setting explicitly, treating
classification as a multi-step reasoning task with tool use. HSCodeComp%
~\cite{yang2025hscodecomp} evaluates fourteen foundation models, six
open-source agent frameworks instantiated with GPT-5 backbones, and three
closed-source deep-search agents; the strongest baseline reaches 46.8\%
ten-digit accuracy on the HSCodeComp test set, well below the 95\% accuracy
reported for experienced human experts, with the gap not closing under
test-time scaling. Collectively, these three families differ markedly in
their assumptions about where the difficulty of HS classification lies---in
representation learning, in parametric knowledge, or in stepwise
rule application---and no consensus architecture has emerged.

\subsection{Retrieval-augmented LLMs and reasoning over regulated corpora}

Hybrid retrieval that combines lexical scoring with dense neural retrieval
is now standard in production information-retrieval pipelines. ColBERT~%
\cite{khattab2020colbert} introduced fine-grained dense retrieval via late
interaction, and the family has continued to evolve; reciprocal rank
fusion~\cite{cormack2009rrf} provides a calibration-free method for
combining rankings from heterogeneous retrievers and has become a default
choice in hybrid search systems. On the reranking side, listwise LLM
rerankers introduced by RankGPT~\cite{sun2023rankgpt} demonstrated that
permutation-generation prompts over candidate passages with a sliding-window
strategy can outperform supervised cross-encoder rerankers in zero-shot
regimes; subsequent open-source variants have established this as a
standard component of retrieval-augmented pipelines.

Retrieval-augmented reasoning over regulated corpora, including legal and
medical materials, is a closely related research area. Of particular
relevance to our work, Magesh et al.~\cite{magesh2025hallucination}, in a
preregistered empirical evaluation published in the \emph{Journal of
Empirical Legal Studies}, examine commercial legal-research products that
advertise themselves as ``hallucination-free'' through retrieval-augmented
generation. They find that Lexis+ AI hallucinates on approximately 17\% of
queries and Westlaw AI-Assisted Research on approximately 33\%, even in a
setting where retrieval is technically functional. The authors introduce a
two-axis taxonomy that distinguishes \emph{correctness} (whether the
answer is factually right) from \emph{groundedness} (whether the cited
authority actually supports the claim made). This distinction maps cleanly
onto our setting: an HS prediction can be wrong, or it can be right by
accident in the sense that the predicted code is correct while the cited
rule does not actually support it. The latter mode is largely invisible
without structured citation output, and we will return to this distinction
when motivating the design of our final-scoring stage.

\subsection{Agentic LLM workflows: from self-planning to deterministic
pipelines}

The agentic-LLM literature can be read along an axis of how much planning
autonomy is granted to the model. At the autonomous end, ReAct~%
\cite{yao2023react} interleaves reasoning ``thoughts'' with environment
``actions'' in an open-ended loop and has become the canonical reference for
tool-using LLM agents. Reflexion~\cite{shinn2023reflexion} augments such
agents with verbal self-reflection over previous attempts, stored as
episodic memory, to support iterative self-improvement. A large family of
subsequent frameworks (AutoGen, LangGraph, SmolAgents, and others, surveyed
in recent literature) has explored variations on this autonomy theme.

At the lower-autonomy end of the same axis sit \emph{workflow}-style
designs in which the stages are fixed and the LLM is invoked only within
narrow, well-scoped sub-tasks. Recent practitioner-facing discussions have
sharpened the distinction between ``agents'' (autonomously orchestrating a
plan to achieve a goal) and ``workflows'' (executing a predefined sequence
of LLM calls), arguing that for tasks whose stage structure is stable and
known in advance, the workflow style offers lower variance, easier
debugging, and tighter cost control without sacrificing the use of LLMs as
reasoning components. The system reported here sits at the lower-autonomy
end: the stages are dictated by the structure of the HS tariff (chapter,
heading, subheading) rather than discovered at run time, and reflection,
verification, and recovery are retained as local mechanisms within
individual stages rather than as global planning loops. We adopt the term
\emph{deterministic agentic workflow} to make this position explicit, and
the design's implications for accuracy, interpretability, and operational
predictability are discussed in Section~IV.

\section{Method}

\begin{figure*}[t]
\centering
\includegraphics[width=0.92\textwidth]{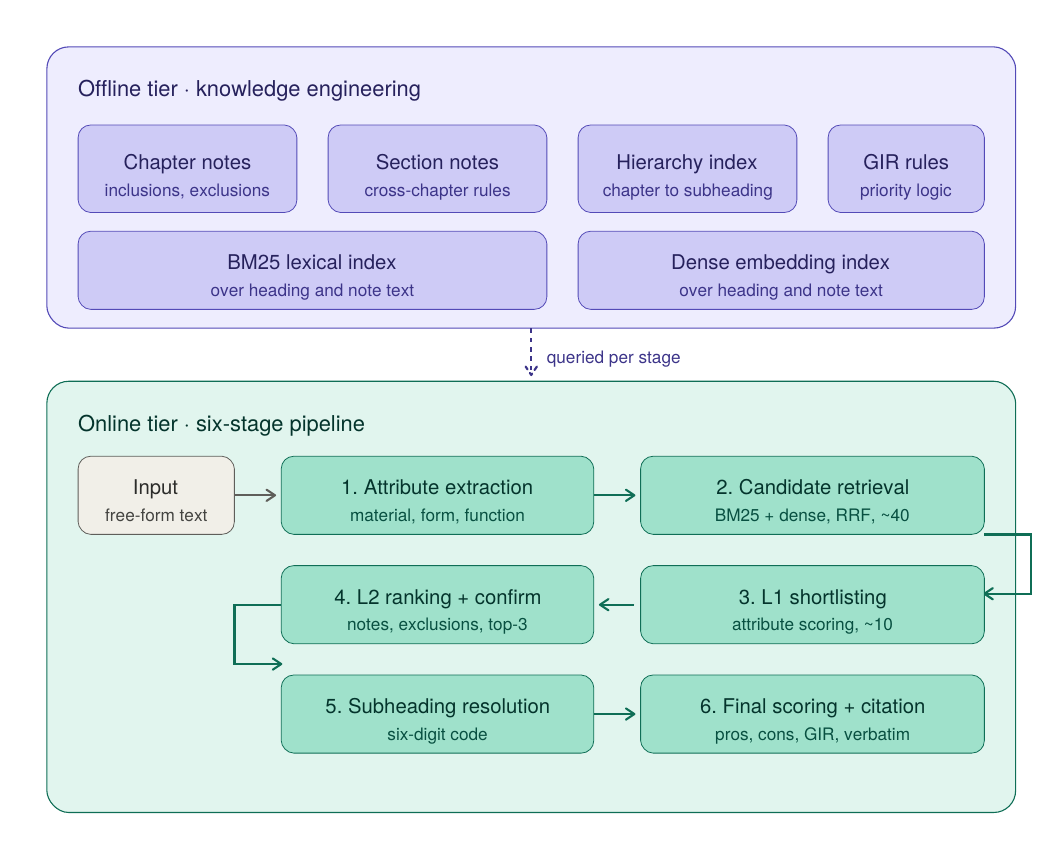}
\caption{System architecture. The offline tier (top) decomposes the
Chinese HS tariff into structured artifacts: chapter notes, section notes,
the chapter-to-subheading hierarchy index, the General Interpretive Rules,
and parallel BM25 and dense embedding indices over heading and note text.
The online tier (bottom) is a six-stage pipeline that queries the offline
artifacts at each stage. Candidate counts narrow from approximately forty
retrieved headings (Stage~2) to ten four-digit candidates (Stage~3) to a
shortlist of three (Stage~4), with subheading resolution and cited final
scoring on top.}
\label{fig:arch}
\end{figure*}

This section describes the system. Section~III-A states the design
philosophy that shapes everything that follows. Section~III-B describes
the offline knowledge-engineering tier that structures the Chinese HS
tariff into queryable artifacts. Section~III-C walks through the online
six-stage classification pipeline. Figure~\ref{fig:arch} gives the overall
architecture; we annotate it as we proceed. Section~III-D presents a
worked example---a TPU mobile-phone screen protector---tracing the input
through all six stages.

\subsection{Design philosophy: deterministic agentic workflow}

The control flow of the system is fixed in advance and is not subject to
LLM-side planning at run time. Each stage is a separate model call with a
strict input schema, a strict output schema, and a stage-local prompt; the
output of one stage is the input of the next. Reflection, verification,
and recovery are retained, but as \emph{local} mechanisms within stages.
For instance, the L2 stage discussed below may re-rank or demote a
candidate when a chapter-note exclusion is detected; this is reflection
inside the stage, not a global re-plan of the pipeline.

This is a deliberate counter-position to deep-search agents of the
ReAct--Reflexion lineage. We do not claim that planning autonomy is bad
in general; we claim that for tariff classification, the stage structure
is dictated by the tariff itself (chapter, heading, subheading) and is
not something the model needs to discover. Trading planning autonomy for
determinism in this setting buys three concrete properties:
interpretability by construction (every stage emits an inspectable
artifact), variance reduction (the same input traverses the same stages
in the same order on every call), and operational legibility (failures
are localizable to specific stages without re-running the pipeline).

\subsection{Offline tier: HS tariff knowledge engineering}

The Chinese HS tariff is published as a structured document with section
notes (one per section, of which there are 21), chapter notes (one or
more per chapter, of which there are 99), explanatory notes from the WCO
adapted by national customs, and the heading and subheading texts
themselves. We decompose this corpus offline into five families of
artifacts.

\emph{Note artifacts.} Each section note and chapter note is parsed into
typed clauses: \emph{inclusion clauses} that admit named goods into a
chapter, \emph{exclusion clauses} that explicitly route named goods out of
the chapter (typically into a named alternative heading), and
\emph{priority clauses} that establish form-priority or part-versus-whole
disciplines internal to the chapter. The chapter-note exclusions are
particularly load-bearing: many of them encode cross-chapter redirections
that an LLM would not otherwise see in a single retrieval pass.

\emph{Hierarchy index.} The tariff hierarchy is materialized as a
queryable index from chapter to heading to subheading, with each level
linked to its applicable notes. Given a candidate heading, the system can
retrieve in $O(1)$ time the full set of notes that constrain its
classification.

\emph{GIR rules.} The six General Interpretive Rules are encoded as
priority statements that the online stages can invoke explicitly. GIR~1
(headings, then notes), GIR~3(a) (specific over generic), GIR~3(b)
(essential character for composite goods), and GIR~6 (subheading-level
analogues of GIR~1 and GIR~3) carry most of the weight in practice.

\emph{Retrieval indices.} Two parallel indices are built over heading and
note text. A BM25 index handles lexical matches on rare and concrete
nouns (``polyurethane'', ``centrifugal'', ``thermosetting''). A dense
embedding index handles functional paraphrases (``for vehicle cooling'',
``anti-corrosion coating''). They are combined at retrieval time by
Reciprocal Rank Fusion~\cite{cormack2009rrf}, chosen because it requires
no score calibration between the two retrievers.

The offline tier is built once per tariff version. When the tariff is
revised (in China typically annually, with mid-year notices), only the
affected chapters and headings are re-indexed.

\subsection{Online tier: six-stage classification pipeline}

\textbf{Stage~1 (attribute extraction)} converts the free-form product
description into a structured set of attributes: material, form and
processing state, function and intended end-use, salient distinguishing
features, and---where reliably inferable from the description---origin or
brand-name disambiguation. This externalizes the implicit pre-reading
that a human classifier performs before opening the tariff. The schema
is fixed; the model is not asked to invent attribute keys.

\textbf{Stage~2 (candidate retrieval)} queries the BM25 and dense indices
in parallel using both the original description and the Stage~1
attributes, and fuses the two ranked lists by RRF. The output is roughly
forty heading candidates. Recall, not precision, is the goal here; later
stages discriminate. Tuning the BM25/dense weighting by hand was
consistently less productive than refining the prompts of the subsequent
ranker stages, and we therefore retain the unweighted RRF fusion.

\textbf{Stage~3 (L1 shortlisting)} scores each of the forty candidates
against the Stage~1 attributes via an LLM ranker prompted with the
heading text alone (no notes yet). The output is approximately ten
four-digit candidates. The intent is to do cheap discrimination before
loading the expensive chapter-note context.

\textbf{Stage~4 (L2 ranking with note-triggered confirmation)} loads the
chapter notes and section notes that bear on the surviving candidates and
performs a second, deeper rank. This stage is given explicit permission
to demote a candidate purely on the strength of an exclusion clause---a
pattern that mirrors expert reasoning, in which an apparently strong
heading is sometimes ruled out by a single note in the chapter to which
the candidate belongs. A note-triggered \emph{confirmation} sub-step then
forces a commitment at the four-digit level, with structured pros, cons,
and the applicable GIR clauses. The output is three four-digit
candidates.

\textbf{Stage~5 (subheading resolution)} descends from the chosen
four-digit heading to its six-digit subheadings, applying GIR~6 (the
subheading analogue of GIR~1 and GIR~3) together with subheading-level
notes. Where the description unambiguously implies a national eight-digit
subdivision, the eight-digit code is also emitted; otherwise the system
stops at six digits and flags the ambiguity.

\textbf{Stage~6 (final scoring with structured citation)} produces the
final output: the top three candidate codes, each annotated with a
confidence score, supporting points (pro), opposing points (con), the
GIR clause(s) applied, and \emph{verbatim} citations from the chapter or
section notes that bear on the decision. Citation strings are
post-processed to annotate which subset of the cited rule was
load-bearing. The two-axis evaluability that this output schema enables
is discussed in Section~IV; the structure also maps directly onto the
correctness--versus--groundedness distinction of
Magesh et al.~\cite{magesh2025hallucination}.

\subsection{A worked example}

To make the pipeline concrete, consider a typical small-importer
description: \emph{``50 pcs HD hydrogel film for all phones, matte /
anti-blue-ray / UV / privacy, TPU material.''} This is a real entry from
the HSCodeComp benchmark. Stage~1 extracts \texttt{material =
thermoplastic polyurethane (TPU)}, \texttt{form = self-adhesive thin
film}, \texttt{function = screen protection}, \texttt{end-use = mobile
phone screen}, and notes the marketing terms (matte, anti-blue-ray) as
non-classifying decorations. Stage~2 retrieves a mix of plastic-film
headings (39.19, 39.20, 39.26) and electronics-related headings
(85.17 parts).

At Stage~3, the L1 ranker reads the heading texts alone and shortlists
39.19 (``self-adhesive plates, sheets, film, and other flat shapes'',
plastics), 39.20 (``other plates, sheets, film, plastics''), and 39.26
(``other articles of plastics''), among others. At Stage~4, loading the
Chapter~39 chapter notes makes the form priority explicit: GIR~3(a) and
the named specificity of heading~39.19 establish that any
\emph{self-adhesive} plastic flat shape goes to 39.19 ahead of the
generic 39.20 and the residual 39.26. The confirmation sub-step records
this as the load-bearing clause and commits to 39.19. Stage~5 descends
to subheading~3919.90 (``other'', as the film is not in rolls of width
$\leq$~20\,cm, which would have indicated 3919.10). Stage~6 emits the
final scored output with the verbatim text of heading~39.19, the GIR~3(a)
specificity clause, and the chapter-note text that confirms
self-adhesion is sufficient for 39.19 regardless of any printed pattern.

The example also illustrates a failure mode that the pipeline is
designed to prevent. A model prompted end-to-end on the same description
will often land on 3926.90, ``other articles of plastics''. The
underlying mistake is that the model resolves the material axis
correctly (plastic) but does not apply the form-priority architecture of
Chapter~39---which is precisely the constraint loaded by Stage~4.

\section{Experiments}

\subsection{Setup}

\textbf{Benchmark.} We evaluate on
HSCodeComp~\cite{yang2025hscodecomp}, an expert-level dataset of 632
product entries collected from a large-scale e-commerce platform and
annotated with ten-digit HS codes by multiple e-commerce domain experts.
Although HSCodeComp's ten-digit codes follow the U.S.\ Harmonized Tariff
Schedule, the upper levels of the hierarchy (chapter, four-digit
heading, six-digit subheading) are part of the international Harmonized
System and are directly applicable to the Chinese regime. We therefore
report results at the four-digit and six-digit levels, and treat the
ten-digit level as outside the scope of a Chinese-tariff system.

\textbf{Backbones.} We evaluate two backbones: \emph{Qwen3.6-plus}, a
frontier proprietary model accessed via API, used in non-thinking mode;
and \emph{Qwen3.6-27B-FP8}, an open-weight 27B-class model in FP8
quantization, deployed locally, also in non-thinking mode. Both backbones
are used through the same workflow; no per-backbone prompt tuning is
performed. The intent is to test whether the workflow itself, rather than
a particular backbone's reasoning capacity, drives the result.

\textbf{Metrics.} We report top-1 and top-3 accuracy at the four-digit
and six-digit levels: a prediction is correct if any of the top-3
candidate codes emitted by Stage~6 matches the ground-truth code at the
relevant granularity. We additionally report cross-backbone
\emph{agreement}: for the same input, the fraction of cases on which
Qwen3.6-27B-FP8 emits the same top-1 prediction as Qwen3.6-plus. Stage
attribution counts the number of cases in which the gold heading is
discarded at each pipeline stage.

\textbf{Versioning.} The system has been iterated over multiple
production releases. We report results for the v7 release, which
introduced expanded recall vocabulary in Stage~2, L1 retention of
note-tagged candidates, a self-exclusion rule in Stage~6 ranking, and
broadened triggers in the L2 confirmation sub-step.

\subsection{End-to-end accuracy}

Table~\ref{tab:e2e} reports end-to-end accuracy on the 632-sample
HSCodeComp evaluation. With Qwen3.6-plus as the backbone, the workflow
reaches 75.0\% top-1 and 91.5\% top-3 at the four-digit level, and
64.2\% top-1 and 78.3\% top-3 at the six-digit level. The four-digit
top-3 figure, in particular, indicates that for over nine out of ten
queries the gold heading is contained in the system's emitted shortlist.
The gap between top-1 and top-3 at four digits (75.0\% to 91.5\%, i.e.,
16.5~pp) suggests that the residual error at top-1 is concentrated in
borderline four-digit decisions where multiple headings are individually
defensible, an observation we revisit in the manual audit below.

\begin{table}[t]
\caption{End-to-end accuracy on HSCodeComp ($n{=}632$).}
\label{tab:e2e}
\centering
\renewcommand{\arraystretch}{1.15}
\begin{tabular}{lcc}
\toprule
Metric & top-1 & top-3 \\
\midrule
Four-digit accuracy & 75.0\% & 91.5\% \\
Six-digit accuracy  & 64.2\% & 78.3\% \\
\bottomrule
\end{tabular}
\end{table}

\subsection{Stage attribution of four-digit errors}

For each query whose top-1 four-digit code does not match the gold
heading, we trace the pipeline output to identify the stage at which the
gold heading was lost or down-weighted. Table~\ref{tab:attr} summarizes
the result for the 158 four-digit top-1 errors observed under v7.

\begin{table}[t]
\caption{Stage attribution of four-digit top-1 errors (158 cases).}
\label{tab:attr}
\centering
\renewcommand{\arraystretch}{1.15}
\begin{tabular}{lr}
\toprule
Stage at which the gold heading is lost & Cases \\
\midrule
Recall miss (Stage 2)                                & 35 \\
L1 drop (Stage 3)                                    & 3  \\
L2 not in keep\_final (Stage 4)                      & 16 \\
L2 keeps gold but not top-1 (Stage 4)                & 102\\
Final ranking override (Stage 6)                     & 2  \\
\midrule
Total                                                & 158\\
\bottomrule
\end{tabular}
\end{table}

Recall failures account for 22\% of all four-digit errors; in the
overwhelming majority of cases (102 out of 158, 65\%) the gold heading
is retrieved, retained through L1, and kept by L2, but not placed first
at the L2 ranking step. This is the failure mode the L2 stage is most
exposed to: the gold heading is correctly recognized as a candidate, but
the ranker prefers a sibling on grounds that, in retrospect, miss a
priority constraint such as form priority or specific-over-generic.

\subsection{Backbone substitutability}

Table~\ref{tab:agree} reports the prediction agreement between the
Qwen3.6-27B-FP8 backbone and Qwen3.6-plus on the same 632 inputs under
the same workflow. For top-1 predictions, the two backbones agree on
84.2\% of four-digit codes and 77.4\% of six-digit codes.

\begin{table}[t]
\caption{Top-1 prediction agreement between Qwen3.6-27B-FP8 and
Qwen3.6-plus on the same 632 inputs and the same workflow.}
\label{tab:agree}
\centering
\renewcommand{\arraystretch}{1.15}
\begin{tabular}{lc}
\toprule
Level & Agreement \\
\midrule
Top-1 four-digit & 84.2\% (532/632) \\
Top-1 six-digit  & 77.4\% (489/632) \\
\bottomrule
\end{tabular}
\end{table}

The result is consistent with the design hypothesis: when the workflow
fixes the control flow and curates the context entering each LLM call,
the choice of backbone within a reasonable capability tier becomes less
load-bearing than it is for end-to-end prompting. A smaller open-weight
27B-class backbone in FP8, run in non-thinking mode on modest hardware,
produces predictions that align with the frontier model on roughly four
out of five four-digit decisions, indicating the workflow can be operated
without dependence on a frontier proprietary model.

\subsection{Manual audit of disagreements}

The accuracy numbers in Table~\ref{tab:e2e} compare predictions against
HSCodeComp's published ground-truth labels and assume those labels are
correct. To examine that assumption, we manually audit the 226 six-digit
top-1 disagreements between Qwen3.6-plus predictions and HSCodeComp
ground-truth.

The audit follows a two-stage protocol. First, each disagreement is
assessed by an external large language model (distinct from the
classification backbone) prompted with the HS Explanatory Note text, the
applicable chapter and section notes, and a strict instruction to base
its judgment only on the rule text rather than on case analogy. Second,
each disagreement is reviewed by a human expert with customs-classifica\-tion
training, who confirms or overrides the model's judgment. The expert
also assigns each disagreement to one of four buckets:
(A)~the agent's prediction appears better supported by the rule text;
(B)~the published ground truth appears better supported (the agent is
wrong); (C)~the case is a genuine HS boundary where professional
classifiers could reasonably disagree; or (D)~the query text or label
exhibits data-quality issues (e.g., the product description and the
ground-truth code are mutually inconsistent).

Table~\ref{tab:audit} summarizes the bucket distribution. We emphasize
that the audit is preliminary. Annotation noise and the inherent
subjectivity of boundary cases mean that the absolute proportions should
not be taken as definitive; the most defensible reading is that the
proportions in buckets~A and~D are large enough that headline accuracy
numbers computed against the unaudited labels are likely to underestimate
the true performance.

\begin{table}[t]
\caption{Manual audit of 226 six-digit disagreements.}
\label{tab:audit}
\centering
\renewcommand{\arraystretch}{1.15}
\begin{tabular}{p{0.58\columnwidth}rr}
\toprule
Bucket & Cases & Share \\
\midrule
A. Agent better supported by HS rules & 96  & 42.5\% \\
B. Ground truth better supported (agent wrong) & 34  & 15.0\% \\
C. HS boundary, professional disagreement plausible & 83  & 36.7\% \\
D. Data-quality issue in query or label & 13  & 5.8\%  \\
\midrule
Total & 226 & 100\% \\
\bottomrule
\end{tabular}
\end{table}

Treating bucket~A and bucket~D as ``agent-not-incorrect''
yields a corrected six-digit top-1 accuracy of approximately 85.8\% on
this benchmark, with the caveat above. A more aggressive accounting that
also credits boundary cases (bucket~C) yields approximately 96.2\%; we
report the conservative figure as the one we are willing to defend
without case-by-case verification by a second expert.

\subsection{Failure mode analysis}

Among the 34 bucket-B cases (cases on which the system is, on audit,
genuinely wrong), we identify five recurring patterns:
(i)~semantic confusion in Chapter~71 between cubic zirconia, rhinestone,
synthetic stones, and reconstructed stones;
(ii)~boundary between watch-case bodies (9111.x) and other clock or
watch parts (9114.x) in Chapter~91;
(iii)~insufficient surfacing of form priority in Chapter~39 at the L2
stage, so that material-named subheadings (e.g., 3907.70 for PLA)
occasionally beat form-named subheadings (e.g., 39.16 for monofilaments
above 1\,mm);
(iv)~parts mistakenly classified to a generic parts chapter rather than
to the functional chapter of the parent machine, in violation of
Section~XVI and Section~XVII Note~2;
(v)~inconsistent application of GIR~3(a) specific-over-generic when the
candidate set contains a residual basket heading.
The five patterns suggest where the offline knowledge base, in
particular the encoding of chapter-internal priority architectures and
the cross-chapter exclusion graph, can be strengthened in subsequent
releases.

\section{Conclusion}

We presented a deterministic agentic workflow for HS tariff
classification. The architectural commitment is to fix the control flow
in advance, confine the LLM to narrow stages with structured inputs and
outputs, and treat reflection and verification as stage-local mechanisms
rather than global planning loops. The system combines an offline
knowledge-engineering tier that decomposes the Chinese HS tariff into
typed rule artifacts with an online six-stage pipeline that traverses the
tariff hierarchy and emits decisions with verbatim rule citations.

On HSCodeComp, the workflow reaches 75.0\% top-1 and 91.5\% top-3
four-digit accuracy and 64.2\% top-1 and 78.3\% top-3 six-digit accuracy
with a frontier backbone; an open-weight 27B-class backbone in
non-thinking mode produces top-1 predictions that agree with the frontier
backbone on 84.2\% of four-digit and 77.4\% of six-digit cases, indicating
that the workflow does not depend on the reasoning capacity of any single
state-of-the-art model. A manual audit of 226 six-digit disagreements,
following an external-LLM-assisted, expert-reviewed protocol, suggests
that a non-trivial fraction of the published ground-truth labels on this
benchmark may deviate from HS general rules; we report this finding as
preliminary and release the adjudication record for community review.

Several directions remain open. We are fine-tuning a 9B-class model on
the intermediate artifacts produced by the workflow; the training is
ongoing and early results are encouraging, to be reported separately.
The cross-chapter exclusion graph implied by the chapter notes is
currently exploited heuristically through Stage~4 retrieval; a more
systematic graph-based treatment is a natural next step. Finally, the
overall template---a deterministic workflow over a hierarchically
organized regulatory corpus, with structured rule-cited
output---generalizes to adjacent tasks such as sanitary classification,
dual-use export control, and drug schedule classification, where the
answer is constrained by a fixed corpus and the reasoning must be
auditable.

\onecolumn
\raggedbottom
\newpage

\noindent{\large\textbf{Appendix: Manual audit of 226 six-digit disagreements}}

\vspace{0.6em}

\noindent
The table below shows 26 representative cases from the manual audit of
226 six-digit top-1 disagreements between system predictions and
HSCodeComp ground-truth labels (Section~IV-E). Columns: HSCodeComp task
ID; product; published ground-truth six-digit code (GT); system
prediction (Agent); expert-assigned bucket; one-line rule-based
rationale citing HS Explanatory Notes or chapter or section notes.
Buckets: \textbf{A}---system better supported by HS rules than the
published GT (GT may deviate); \textbf{B}---published GT better
supported, system is wrong; \textbf{C}---genuine HS boundary;
\textbf{D}---data-quality issue in query or label. The complete record,
per-case citations, and full nine-stage traces are available from the
corresponding author on request.

\vspace{0.6em}

\noindent
\scriptsize
\renewcommand{\arraystretch}{1.08}
\begin{tabular}{@{}p{0.025\textwidth}p{0.20\textwidth}p{0.06\textwidth}p{0.06\textwidth}p{0.025\textwidth}p{0.54\textwidth}@{}}
\toprule
\textbf{ID} & \textbf{Product} & \textbf{GT} & \textbf{Agent} & \textbf{Bk} & \textbf{Rule-based rationale} \\
\midrule
\multicolumn{6}{l}{\textit{Bucket A --- system's prediction better supported by HS rules (GT may deviate)}} \\
\midrule
350 & TPU self-adhesive screen film & 3926.90 & 3919.90 & A & Heading 39.19 names self-adhesive plastic flat shapes specifically; GIR~3(a) places it ahead of the residual basket 39.26. \\
287 & ANNKE 8CH network video recorder & 8522.90 & 8521.90 & A & A network video recorder is a complete recording apparatus directly listed in 85.21; it is not a part within the scope of 85.22. \\
7   & Zinc-alloy with rhinestone jewelry set & 7116.20 & 7117.19 & A & Rhinestones are glass imitations of gemstones (not natural, synthetic, or reconstructed stones per Chapter 71 Note 1); jewelry with glass imitations falls in 71.17. \\
608 & 99.8\% high-concentration tattoo anesthetic cream & 3304.99 & 3004.90 & A & High-concentration anesthetic for permanent makeup is a medicament; 30.04 names this use specifically, ahead of cosmetic heading 33.04. \\
316 & Untreated natural human hair, bulk & 6704.20 & 0501.00 & A & Heading 05.01 covers human hair, unworked; 67.04 covers wigs and false eyelashes (worked articles), to which the product has not yet been processed. \\
391 & Transparent PVC ankle rain boots & 6402.19 & 6401.92 & A & Heading 64.01 specifically names waterproof footwear of rubber or plastics; PVC waterproof boots are therefore precluded from 64.02. \\
117 & Nike Air Max running shoes & 6404.19 & 6404.11 & A & Subheading 6404.11 specifically names sports footwear, including training and running shoes; 6404.19 is the residual subheading. \\
558 & Polar-fleece neck gaiter / scarf & 6307.90 & 6117.10 & A & Heading 61.17 specifically names knitted scarves and similar articles; 63.07 is a residual heading for other made-up textile articles. \\
31  & GPS dog-fence wireless collar & 8543.70 & 8526.91 & A & The principal function is radio-navigation positioning, specifically named in 85.26.91, ahead of the residual heading 85.43 for unnamed electrical apparatus. \\
32  & Fractional CO\textsubscript{2} laser skin device & 8543.70 & 9018.90 & A & A professional medical / aesthetic laser instrument is more specifically classified in 90.18 than in the residual heading 85.43. \\
472 & Activated-carbon replacement filter cartridge & 8421.21 & 8421.99 & A & A replacement consumable for a filtration apparatus is a part, classified in 8421.99, not in the integrated-apparatus subheading 8421.21. \\
46  & Xiaomi Redmi K40 OLED display assembly & 8517.79 & 8524.92 & A & The 2022 HS introduced 85.24 to specifically cover flat panel display modules; an OLED display assembly falls under 8524.92, more specific than 8517.79 (parts of phones). \\
233 & Loose freshwater pearls for DIY & 7116.10 & 7101.22 & A & Loose pearls not strung into jewelry are classified in 71.01; heading 71.16 covers articles of pearls (i.e., finished jewelry). \\
\midrule
\multicolumn{6}{l}{\textit{Bucket B --- system is wrong; GT is better supported (failure modes identified in Section~IV-F)}} \\
\midrule
137 & PLA 3D-printer filament, 1.75\,mm & 3916.90 & 3907.70 & B & Chapter 39 imposes a form-priority architecture: monofilaments above 1\,mm in cross-section belong to heading 39.16, regardless of the underlying polymer; 3907.70 names PLA in primary form. \\
480 & Car air-conditioning filter element (Mercedes) & 8421.39 & 4823.20 & B & A finished filter element is a complete filtering apparatus and is classified in 84.21 regardless of the substrate material (paper, fabric, or metal); the system mistakenly took the paper-substrate path. \\
52  & Arduino UNO R3 development board & 8542.31 & 8543.70 & B & A general-purpose microcontroller-on-carrier without dedicated end-function is admissible under 85.42 as a packaged assembly of integrated circuits; the system over-applied the 8542 exclusion clause and landed in the residual 85.43. \\
606 & Stainless-steel cleaning filter basket (wire mesh) & 7326.90 & 7326.20 & B & Subheading 7326.20 specifically names articles of iron or steel wire, to which a wire-mesh basket conforms; the system selected the residual 7326.90 unnecessarily. \\
569 & Anti-theft bicycle padlock (alloy steel) & 8301.40 & 8301.10 & B & Subheading 8301.10 specifically names padlocks; 8301.40 covers other locks. \\
\midrule
\multicolumn{6}{l}{\textit{Bucket C --- HS boundary, professional classifiers may disagree}} \\
\midrule
20  & DIATONE BLHeli\_32 brushless ESC & 8503.00 & 8504.40 & C & An electronic speed controller for a brushless motor sits between motor-part heading 85.03 and static-converter heading 85.04; both are defensible in current practice. \\
295 & LG washing-machine MCU control board & 8537.10 & 8450.90 & C & The application of Section~XVI Note~2(a) vs.\ Note~2(b) is contested for MCU-bearing appliance control boards; both 85.37 and 8450.90 have precedent. \\
113 & Halter-neck strappy bodysuit (polyester) & 6211.43 & 6204.33 & C & Heading 62.11 (casual / sportswear) and heading 62.04 (women's upper garments) overlap for halter-neck strappy tops; national practice varies. \\
274 & UGREEN HDMI-compatible video cable & 8544.20 & 8544.42 & C & HDMI cables are technically shielded multi-conductor, between coaxial-cable 8544.20 and 8544.42 (other conductors with connectors); current administrative practice differs by country. \\
\midrule
\multicolumn{6}{l}{\textit{Bucket D --- data-quality issue in query or label}} \\
\midrule
108 & ``Korean luxury denim distressed jeans'' & 6103.42 & 6203.42 & D & The query describes denim trousers (woven by construction); the GT records a knitted subheading (6103), inconsistent with the product. \\
543 & Confederate flag, 100D polyester, 3$\times$5\,ft & 4911.91 & 6304.93 & D & The query self-annotates that the item is a printed-and-dyed polyester flag; the GT codes it as printed matter (49.11), inconsistent with textile flag products. \\
627 & USB car charger with cigarette-lighter plug & 9613.80 & 8504.40 & D & The GT records 96.13 (lighters), incompatible with the product, which is a static converter for in-vehicle charging. \\
\bottomrule
\end{tabular}

\end{document}